\documentclass[11pt]{article}

\usepackage{geometry}
\usepackage{setspace}
\usepackage{caption} 
\usepackage[table]{xcolor}
\usepackage{graphicx}
\usepackage{url}
\usepackage{multirow}

\usepackage{mathptmx}
\usepackage[T1]{fontenc}

\usepackage{natbib}
\setcitestyle{aysep={}}

\newenvironment{userabstract}{%
\begin{quote} \bf}
{\end{quote}}

\newenvironment{userkeywords}{%
\begin{quote} \it \textbf{Keywords:}}
{\end{quote}}

\newenvironment{useracknowledgements}{%
  \footnotesize  \textbf{Acknowledgements:}}

\title{Discovering Directly-Follows Graph Model for Acyclic Processes} 

\author
{Nikita Shaimov$^{1\ast}$, Irina Lomazova$^{1}$, Alexey Mitsyuk$^{1}$\\
\\
\normalsize{$^{1}$Faculty of Computer Science (PAIS Lab), HSE University, Moscow, Russia}\\
\\
\normalsize{$^\ast$Corresponding author: Nikita Shaimov, E-mail: nshaimov@hse.ru}
}

\date{}

\usepackage{amsthm, amssymb}
\theoremstyle{definition}
\newtheorem{definition}{Definition}
\usepackage{tikz-cd}

\begin{document}

\baselineskip18pt

\maketitle 

\begin{userabstract}

Process mining is the common name for a range of methods and approaches aimed at analysing and improving processes. Specifically, methods that aim to derive process models from event logs fall under the category of process discovery. Within the range of processes, acyclic processes form a distinct category. In such processes, previously performed actions are not repeated, forming chains of unique actions. However, due to differences in the order of actions, existing process discovery methods can provide models containing cycles even if a process is acyclic. This paper presents a new process discovery algorithm that allows to discover acyclic DFG models for acyclic processes. A model is discovered by partitioning an event log into parts that provide acyclic DFG models and merging them while avoiding the formation of cycles. The resulting algorithm was tested both on real-life and artificial event logs. Absence of cycles improves model visual clarity and precision, also allowing to apply cycle-sensitive methods or visualisations to the model.

\end{userabstract}

\begin{userkeywords}
Process mining, Process models, Process discovery, Directly-follows graphs, Acyclic graphs, Event logs
\end{userkeywords}

\begin{useracknowledgements}
This research was supported by the Basic Research Program of the National Research University Higher School of Economics.
\end{useracknowledgements}

%------------------------------------------------------------------------------
\section{Introduction}\label{sec:introduction}

For the purpose of process data analysis, \emph{process mining} \citep{Aalst16} methods have been successfully applied in many areas, such as business, medicine, logistics, and many others. Process mining allows us to explore, analyse and improve processes through event data stored in the event logs. The event log is the collection of event records that reflect process behaviour for a certain period of time. 
Process mining methods can be divided into three categories: \textit{process discovery}, \textit{conformance checking}, and \textit{process enhancement}. 
\textit{Process discovery} methods are very important \citep{guide} as they allow for the automatic synthesis of models of real processes based on event logs. 
Discovered process models can be analysed in order to search for anomalies, correlations, deviations and inefficiencies, taking into account the flow of the process. 
Depending on the algorithm, models can be synthesised in various notations, such as Petri nets, UML, BPMN or Directly Follows Graphs (DFGs). 
Despite its drawbacks \citep{vanderAalst2020}, we consider DFG to be one of the most popular notations due to its simplicity and the fact that it does not require specific knowledge of the notation for its interpretation. Our assertion is supported by extensive implementations of the DFG notation in many popular process mining tools such as \emph{PM4Py} \citep{BERTI2023100556}, \emph{Celonis} \citep{Celonis}, and \emph{Disco} \citep{Disco}.

\begin{figure}[htb]
    \centering
    \includegraphics[width=0.5\textwidth]{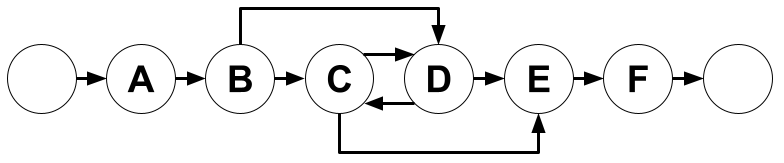}
    \caption{An example DFG with cycles}
    \label{fig:cycle_model}
\end{figure}

Most of the existing process discovery algorithms synthesise models in which each unique action in the process is represented by a single node. This can lead to the appearance of fake cycles in models of intrinsically acyclic processes. 
In this paper, we call a process acyclic if, in any sequence of actions it performs, each action is unique. 
For example, let us take a process with only two possible sequences of actions: $\langle A, B, C, D, E, F \rangle$ and $\langle A, B, D, C, E, F \rangle$. This process is acyclic, as actions in both sequences do not repeat. The only difference between sequences is the order of actions $C$ and $D$. This difference, combined with existing process discovery algorithms, leads to the occurrence of cycles in the resulting DFG model (see Fig.~\ref{fig:cycle_model}).

In addition to complexity, cycles in models of acyclic processes reduce their precision as they allow reproducing non-existent sequences of actions. In the model from Fig.~\ref{fig:cycle_model}, action sequences $\langle A, B, D, E, F \rangle$ or $\langle A, B, C, D, C, E, F \rangle$ can be replayed, but they are not present among the initial action sequences.

Therefore, the aim of this paper is to provide a new process discovery algorithm that constructs an acyclic DFG process model for an acyclic process. 
Our algorithm discovers the model by partitioning the event log into parts that provide acyclic DFG process models. 
Then, it constructs an acyclic DFG model of the whole process by merging these partial acyclic DFG models. In order to avoid the appearance of cycles, we allow several instances of the same action node in the model.

Our paper answers the following \emph{research questions}:
\begin{itemize}
    \item Q1 How to discover an acyclic DFG for an acyclic process?
    \item Q2 How precise is the model discovered by our algorithm compared to the DFG discovered by the standard algorithm?
    \item Q3 Is our discovery algorithm applicable to artificial and real event data?
\end{itemize}

The rest of the paper is organised as follows. In Section~\ref{sec:motivation}, based on the example, we consider the causes and consequences of cycles in process models of acyclic processes. Section~\ref{sec:definitions} provides definitions that will be 
employed in our paper. 
Section~\ref{sec:discovery} presents the new algorithm for discovering acyclic DFGs. 
Section~\ref{sec:evaluation} provides the results of its evaluation on real and artificial event logs. 
Section~\ref{sec:related} discusses related work. Section~\ref{sec:conclusion} concludes the work and provides directions for future work.

%------------------------------------------------------------------------------
\section{Motivating example}\label{sec:motivation}

Let us begin with a real-life example to demonstrate the realistic nature of the problem we are studying.
In this paper, we consider data from the Learning Management System (LMS) of our university. 
The data contains information about students' academic performance. Based on these data, we create an event log.
In this paper, the event log contains  data on events related to  three groups of students  attending classes, lectures, and seminars in two academic courses: ``Business and management in a global context'' (``bmgc'', for short) and ``Calculus'' (``calc'', for short). 
The log consists of traces, which are sequences of actions ordered by time.
Each trace is related to a single case, which in our context corresponds to a student. 
Thus, the collection of traces represents individual learning trajectories of students.

We would like to motivate our research using the small simplified sample from the event log. 
Our sample includes records for five students and only parts of events (see Tab.~\ref{tbl:example}). 
Each row represents an event record. 
The record contains the case ID, timestamp, and the name of the  activity performed. 
In our example, the  records have the following form: \texttt{<the\_name\_of\_the\_educational\_course> <the\_type\_of\_the\_performed\_class>\_<the\_serial\_number\_of\_the\_class>}. 
The \texttt{timestamp} shows when the class took place, and the \texttt{case\_ID} is the unique identification number of the student.

\begin{table}[htb]
    \centering
    \caption{The event log example}
    
    \begin{tabular}{c c c} 
    \textbf{Case\_ID} & \textbf{Timestamp} & \textbf{Activity} \\ 
    \hline
    \hline
    4 & 04.09.2023 & calc class\_1\\
    \hline
    5 & 04.09.2023 & calc class\_1\\
    \hline
    1 & 06.09.2023 & bmgc lecture\_1\\
    \hline
    3 & 06.09.2023 & bmgc lecture\_1\\
    \hline
    2 & 06.09.2023 & bmgc lecture\_1\\
    \hline
    4 & 06.09.2023 & bmgc lecture\_1\\
    \hline
    5 & 06.09.2023 & bmgc lecture\_1\\
    \hline
    2 & 08.09.2023 & bmgc seminar\_1\\
    \hline
    4 & 08.09.2023 & bmgc seminar\_1\\
    \hline
    5 & 08.09.2023 & bmgc seminar\_1\\
    \hline
    4 & 11.09.2023 & calc class\_2\\
    \hline
    5 & 11.09.2023 & calc class\_2\\
    \hline
    1 & 13.09.2023 & bmgc lecture\_2\\
    \hline
    1 & 13.09.2023 & bmgc seminar\_1\\
    \hline
    3 & 13.09.2023 & bmgc lecture\_2\\
    \hline
    3 & 13.09.2023 & bmgc seminar\_1\\
    \hline
    2 & 13.09.2023 & bmgc lecture\_2\\
    \hline
    1 & 15.09.2023 & calc class\_1\\
    \hline
    2 & 15.09.2023 & calc class\_1\\
    \hline
    3 & 15.09.2023 & calc class\_1\\
    \hline\hline
    \end{tabular}
  \label{tbl:example}
\end{table}

The process model can be constructed with the help of a process discovery algorithm. 
We will use an algorithm that provides a process model in the form of a DFG \citep{vanderAalst2022}. 
DFG represents a process as a graph.
Nodes of this graph correspond to activities.
If an arc connects two nodes, then the related activities  follow each other directly, at least in some instances of the process. 
Thus, such a directed graph reflects all possible orders of process activities observed in reality.

\begin{figure}[htb]
    \centering
    \includegraphics[width=0.35\textwidth]{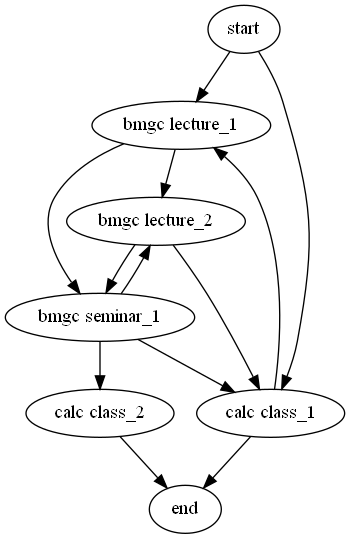}
    \caption{The DFG corresponding to the log shown in Table~\ref{tbl:example}}
    \label{fig:dfg_example}
\end{figure}

The DFG for our sample event log is shown in Fig.~\ref{fig:dfg_example}.
It contains several cycles passing through the following nodes: \{"bmgc lecture\_1", "bmgc lecture\_2", "calc class\_1"\}, \{"bmgc lecture\_2", "bmgc seminar\_1"\}, \{"bmgc lecture\_1", "bmgc lecture\_2", "bmgc seminar\_1", "calc class\_1"\}. 
The presence of cycles reduces the clarity of the model, and, importantly, allows for reproducing nonexistent sequences of events, which reduces the precision of the model. 
For example, sequences $\langle$``bmgc lecture\_1'', ``bmgc lecture\_2'', ``bmgc seminar\_1'', ``bmgc lecture\_2'', ``calc class\_1''$\rangle$ and $\langle$``bmgc lecture\_1'', ``bmgc lecture\_2'', ``calc class\_1''$\rangle$ are not present in the event log.
However, this behaviour can be replayed by the discovered model.

Cycles occur in the DFG of an acyclic process due to different orders of the same activities in process instances.  
Students in the event log (see Tab.~\ref{tbl:example}) have different schedules and therefore attend classes in different orders. 
For example, student with ID 1 attended ``bmgc lecture\_2'' and then ``bmgc seminar\_1'', but student with ID 2 have those classes in reverse order. 
As a result, the model has arcs between nodes ``bmgc lecture\_2'' and ``bmgc seminar\_1'' in opposite directions. 
However, each activity in our process naturally appears only once in any trace of the event log: each student attends each class only once.
This means that the activities are not repeated in real life. 
Therefore, the DFG of such an acyclic process can be misleading.

One way to eliminate spurious cycles from a model is to allow different nodes in the model to be associated with the same activity, i.\ e.
to have several nodes with the same label.
This will increase the size of the model, but it will also increase its precision and eliminate undesirable cycles. In this paper, we describe a new process discovery method that follows this approach and is thus capable of discovering acyclic DFGs for acyclic processes.

%------------------------------------------------------------------------------
\section{Basic definitions}\label{sec:definitions}

In this section, we provide some basic definitions that will be used in the following.

Let $X$ be a set. A \emph{multiset} $m$ over a set $X$ is a mapping: $m : X \rightarrow \mathcal{N}$, where $\mathcal{N}$  is the set of natural numbers (including zero), i.e., a multiset may contain several copies of the same element. 
For an element $x\in X$, we write  $x\in m$ , if $m(x) >0$.
For two multisets $m,m'$ over $X$ we write $m\subseteq m'$  iff $\forall x\in X: m(x) \leq m'(x)$ (the inclusion relation). 
The sum, the union, and the subtraction of two multisets $m$ and $m'$ are defined as usual:
$\forall x\in X: (m+m')(x)=m(x)+m'(x), (m\cup m')(x)=max(m(x),m'(x)), (m-m')(x)=m(x) - m'(x)$, if $m(x) - m'(x) \geq 0$, otherwise $(m-m')(x)= 0$.
By $\mathcal{M}(X)$ we denote the set of all multisets over $X$.

For a set $X$, by $X^*$ with elements of the form $\langle x_1,\dots, x_k \rangle$ we denote the set of all finite sequences (words) over $X$.

In practice, an event log is represented as a collection of records, where each record contains information about an event, including at least the event name, the event identifier, and a timestamp. A collection of records can be divided into traces based on case identifiers, and the order of event names in traces is determined according to the timestamps.

\begin{definition}%[Event log]
Let $E$ be a finite set of \textit{event names} (\textit{events} for short). 
A \textit{trace} over $E$ is a finite sequence $\sigma = \langle e_1, \cdots, e_n \rangle$ of events from $E$, i.e. $\sigma \in E^*$. An \textit{event log} over $E$ is a finite multiset of traces $L \in \mathcal{M}(E^*)$. 
\end{definition}

Given an event log $L$ over $E$, we define two binary relations on $E$ (w.r.t. $L$).

\begin{definition}%[Directly follows ($<$) and follows ($\ll$) relations]
We say that an event $b$ \textit{directly follows} an event $a$ in $L$ (written $a < b$), if there is a trace $\sigma= \langle e_1, \cdots, e_n \rangle\in L$, such that for some $i=1,\cdots, n-1$, $e_i = a$ and $e_{i+1} = b$, i.\ e. $b$ goes directly after $a$ in $\sigma$.

An event $b$ \textit{follows} an event $a$ in $L$, if 
there is a trace $\sigma= \langle e_1, \cdots, e_n \rangle\in L$, such that for some $i,j=1,\cdots, n$, $e_i = a$, $e_j = b$, and $i<j$, i.\ e. $b$ goes  after $a$ in $\sigma$.
\end{definition}

\begin{definition}%[Directly-Follows Graph, DFG]
A \textit{directly-follows graph (DFG)} over $E$ is a labelled directed graph $G = (V, A, \lambda, v_{start}, v_{end})$, where:
\begin{itemize}
    \item $V$ is a finite set of vertices;
    \item $A \subseteq V \times V$ is a finite set of arcs;
    \item $\lambda: V - \{v_{start}, v_{end}\} \rightarrow E$ is a labelling function that maps each  vertex to an event name;
    \item $v_{start} \in V$ is the only vertex that does not have incoming arcs;
    \item $v_{end} \in V$ is the only vertex that does not have outgoing arcs.
\end{itemize}
\end{definition}

DFG is used to model the behaviour of the system. 

\begin{definition}%[Run in a DFG]
Let $G = (V, A, \lambda, v_{start}, v_{end})$ be a DFG over E.
A finite sequence of event names $r = \langle e_1, \cdots, e_n \rangle \in E^*$ is called a run in $G$ if exists a path $\langle v_{start}, v_1, \cdots, v_n, v_{end} \rangle \in G$ with $\lambda(v_1) = e_1,\ \lambda(v_2) = e_2, \cdots,\ \lambda(v_n) = e_n$.
\end{definition}

The following definition relates a DFG with an event log.

\begin{definition}%[Perfect fitness]
We say that DFG $G$ \textit{perfectly fits} event log $L$, if each trace $\sigma \in L$ is a run in $G$.
\end{definition}

Let $L$ be a log in which each trace contains at most one occurrence of each event. We call such a log an \textit{acyclic event log}. In a process represented by such a log, events obviously must not repeat. 

However, as noted earlier, the DFG model discovered from the acyclic log may contain cycles. 
The algorithm for discovering DFG models from event logs in \citep{vanderAalst2022} can be considered as the standard one. So, in our paper, we deal with this algorithm.  
By $DFG(L)$ we denote the DFG  discovered from event log $L$ using this algorithm.

We call an event log $L$ \textit{DFG-acyclic}, if $DFG(L)$ is an acyclic graph.

%------------------------------------------------------------------------------
\section{Acyclic DFG model discovery}\label{sec:discovery}

In this section, we present a new algorithm for discovering an acyclic DFG from an event log of the acyclic process. The main idea of the algorithm is as follows. 
Given an acyclic event log, the first step is to split the log into DFG-acyclic sublogs.
This is always possible because each trace in the acyclic event log can itself be considered as a DFG-acyclic event log.
The classical DFG discovery algorithm is then applied to each of these sublogs, resulting in a set of acyclic DFG (sub)models. 
We then used a special merging of these submodels to construct the acyclic DFG model for the entire original event log.
Note that to avoid cycles in the resulting DFG model, we allow more than one node to be labelled with the same event name.

\begin{figure}[htb]
    \centering
      \begin{tikzcd}[column sep=tiny]
        l_1 \arrow{d} & + & \cdots & + & l_n \arrow{d} & = & L \arrow{d}{} \\
        M_1 & \bigoplus & \cdots & \bigoplus & M_n & = & M
      \end{tikzcd}
    \caption{Compositionality of the ``perfect fitness'' (``$\downarrow$'') relation}
    \label{fig:diagram}
\end{figure}

We illustrate the approach with the commutative diagram in Fig.~\ref{fig:diagram}. 
Here $L$ is an acyclic event log,
 partitioned into DFG-acyclic sublogs $l_1$, $l_2$, $\dots$, $l_n$, i.e.
$L = l_1 +  \dots + l_n$.
Acyclic DFG models $M_1 = DFG(l_1)$, $M_2 = DFG(l_2)$, $\dots$, $M_n = DFG(l_n)$ are discovered from these sublogs using the classical algorithm. By  $\bigoplus$ we denote here special merging operation on submodels $M_1,\dots,M_n$, resulting in a DFG model $M$, corresponding to the entire log $L$. This operation is defined in Sec.~\ref{sec:algo} as a model merging algorithm. 

We show further that $M$ perfectly fits $L$, provided that $M_i$ perfectly fits $l_i$ for all $i=1,\dots,n$.

The following subsections describe the three steps of our discovery algorithm in more detail.
\begin{enumerate}
    \item event log partitioning;
    \item merging of (two) acyclic DFGs;
    \item merging of (two) acyclic DFGs with repeated event names.
\end{enumerate}

%------------------------------------------------------------------------------
\subsection{Acyclic event log partitioning}\label{sec:log_part}
Here we describe how to partition an acyclic event log into DFG-acyclic sublogs.

In an acyclic event log, each event occurs at most once in each trace. For events $A$ and $B$ in an acyclic event log $L$,  we write $A \ll_L B$, or just $A \ll B$, if there is a trace in $L$ with $A$ occurring earlier than $B$.

A DFG model discovered from an acyclic log $L$ contains cycles, if for some two events $A$ and $B$, $A \ll B$ and $B \ll A$ in $L$, i.e., there are two traces in $L$ where two events occur in different orders.
Thus, a single trace is a DFG-acyclic event log. 
Then, a DFG-acyclic sublog should not contain two traces with two events occurring in different orders. We call such traces compatible. Checking whether two traces are compatible is straightforward.

To partition an acyclic event log into DFG-acyclic sublogs, we represent the log as a graph where nodes are traces, and edges connect compatible traces. 
A set of pairwise compatible traces (a DFG-acyclic sublog) forms a clique in this graph. 
Partitioning the log into the minimal number of sublogs is equivalent to finding the minimum clique cover for the graph. 
Since the minimum clique cover problem is NP-hard \citep{Karp1972-la}, we solve the problem using a heuristic algorithm that produces a nearly optimal solution in a reasonable amount of time.

For each of the obtained DFG-acyclic event logs, an acyclic DFG model can be constructed using a known discovery algorithm. 
The next subsection describes how these models can be combined into one overall acyclic DFG model.

%------------------------------------------------------------------------------
\subsection{Merging algorithm for two acyclic DFG models}\label{sec:algo}

After partitioning an acyclic event log and constructing multiple acyclic DFG models, the next task is to combine these models into a new acyclic DFG that perfectly fits the original acyclic event log.  We first describe the merging of two acyclic DFG models with unique event labels. We present two algorithms to solve this problem, a naive one and a more accurate one. However, such a merger may result in a DFG in which multiple nodes are labelled with the same event. Therefore, to merge more than two DFG models, we next present an algorithm to merge acyclic DFG models with duplicate event names.

\subsubsection{Naive Approach}\label{sec:naive}

Now we describe a naive merging algorithm that produces results quickly but at the expense of the size and complexity of the resulting model.

Given two acyclic DFG models with unique event labels, the merging algorithm is performed in four steps:

\begin{enumerate}
    \item searching for maximum common subgraphs;
    \item constructing the connectivity graph for common subgraphs;
    \item finding and eliminating cycles in the connectivity graph;
    \item merging subgraphs based on  fusion of some vertices labelled with the same event names.
\end{enumerate}

The first step is to determine the common part of two DFGs, i.e. to find their maximum common subgraphs. Obviously, in doing so we exclude subgraphs consisting of only one vertex.
In each DFG, each vertex label is unique, so common subgraphs for two DFGs can be obtained by intersecting the arc sets of these DFGs.

\begin{figure}[htb]
    \centering
    \includegraphics[width=0.5\textwidth]{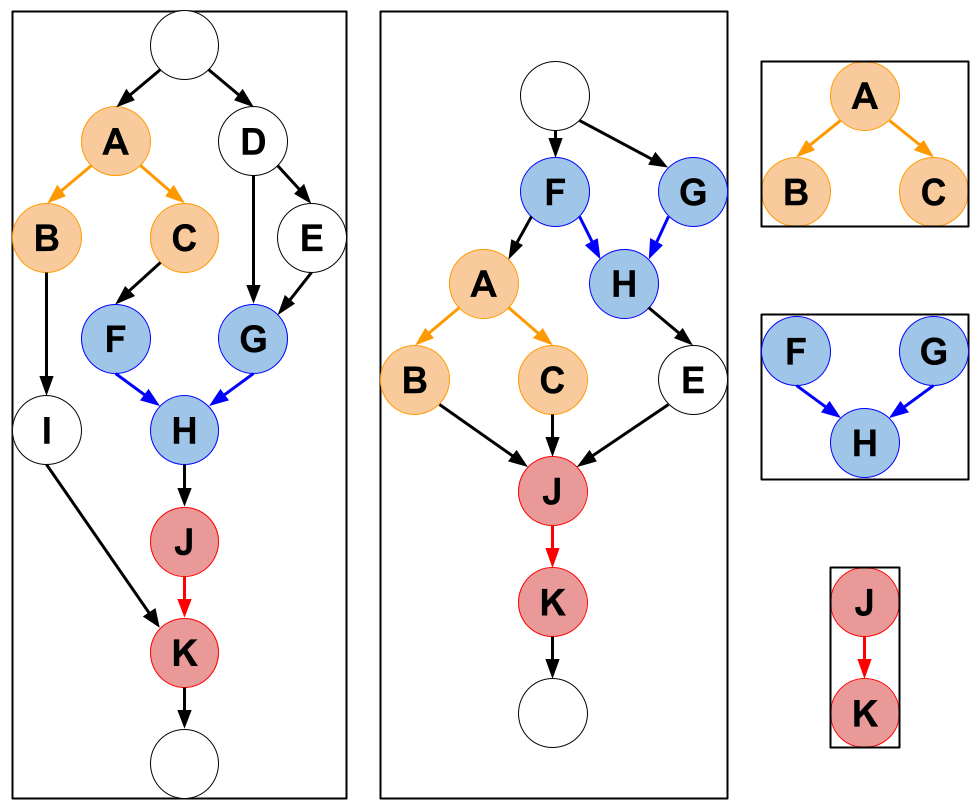}
    \caption{Two DFGs with three common subgraphs: vertices $A$, $B$, and $C$ form the first subgraph; $F$, $G$, and $H$ --- the second one; and vertices $J$, $K$ --- the third one}
    \label{fig:common_subgraphs}
\end{figure}

The following example illustrates the work of the algorithm.
Consider two simple acyclic DFG models shown in Fig.~\ref{fig:common_subgraphs}. 
Vertices related to different common subgraphs are coloured using different shades and are depicted separately in the right part of the figure.

Combining models by merging all common subgraphs may,  generally speaking, introduce cycles in the resulting model. 
This can happen if two common subgraphs precede each other in the two models in different orders, like the yellow and blue subgraphs in Fig.~\ref{fig:common_subgraphs}. 
To determine whether the fusion of subgraph vertices will produce cycles, we construct another directed graph, which we call \emph{connectivity graph}. 
This graph reflects relationships between our common subgraphs; specifically, it shows whether there is a path from some vertex in one subgraph to some vertex in another subgraph. 

\begin{figure}[htb]
    \centering
    \includegraphics[width=0.4\textwidth]{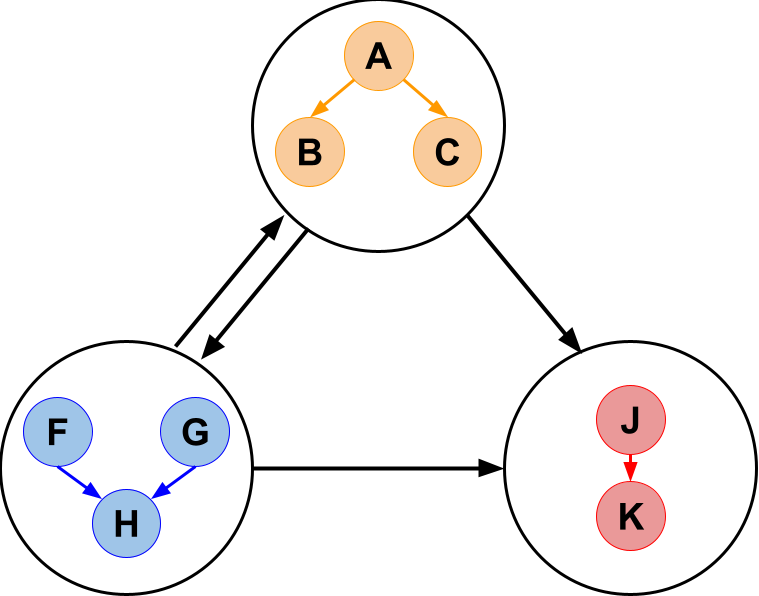}
    \caption{Connectivity graph with three subgraphs as nodes}
    \label{fig:subgraph_as_node}
\end{figure}

In a connectivity graph, nodes are common subgraphs (see the connectivity graph for our example in Fig.~\ref{fig:subgraph_as_node}), and arcs indicate that there is a path from some node in one subgraph to some node in another subgraph in one of the DFG models that need to be merged.

The connectivity graph allows us to determine whether merging common subgraphs produces cycles in the target model. 
The cycles in the connectivity graph indicate the cycles in the merged model. Therefore, we remove some nodes from the connectivity graph to make it acyclic, and do not merge the corresponding common subgraphs, leaving two copies of them in the resulting model.

\begin{figure}[htb]
    \centering
    \includegraphics[width=0.65\textwidth]{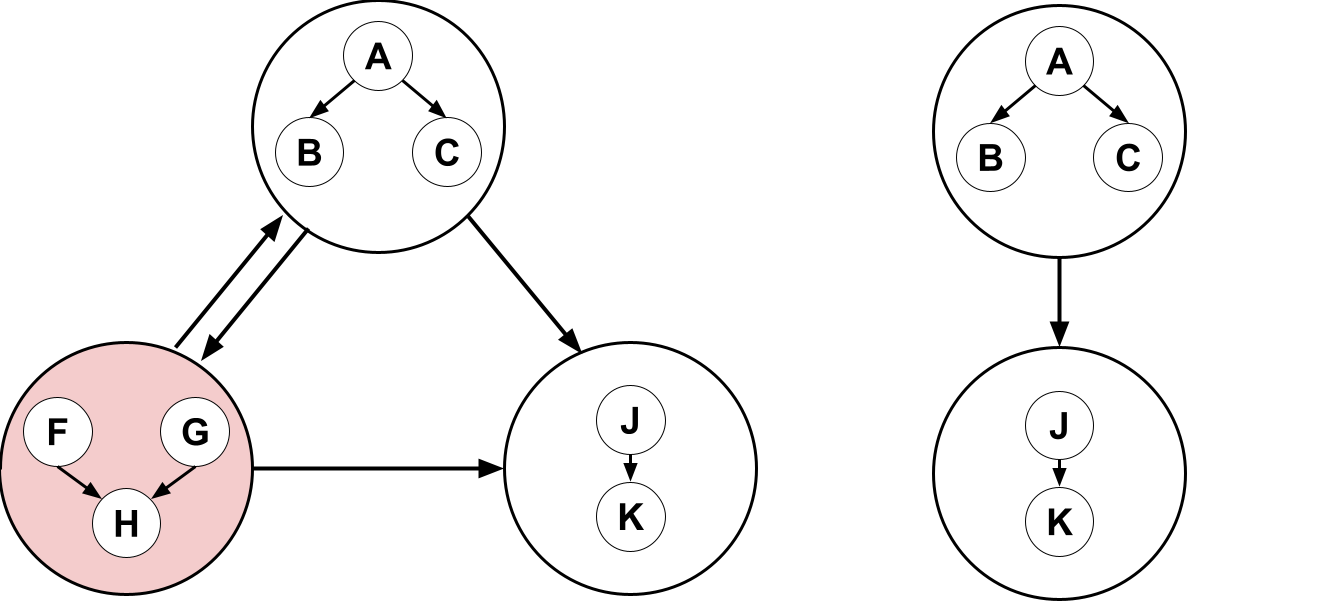}
    \caption{To eliminate the cycle, the vertex $F-G-H$ is excluded from the connectivity graph}
    \label{fig:subgraph_as_node_reduction}
\end{figure}

To get a smaller resulting model, we need to merge as many vertices as possible. For an 
 optimal solution, we need to find the minimum set of vertices in the connectivity graph that we can eliminate to break cycles (see Fig.~\ref{fig:subgraph_as_node_reduction}). 
This problem is known as the minimum feedback vertex set problem. Feedback Vertex Set (FVS) is a set of vertices that, if excluded from a graph, will break all existing cycles. The minimum FVS problem for directed graphs is NP-complete \citep{Karp1972-la}. To solve this problem, we use one of the existing solutions \citep{FeedbackVertexSet}, based on the Bounded Search Tree algorithm \citep{Cygan2015-kt}. It includes a set of graph reductions and recursive branching to determine the minimum FVS.

\begin{figure}[htb]
    \centering
    \includegraphics[width=0.95\textwidth]{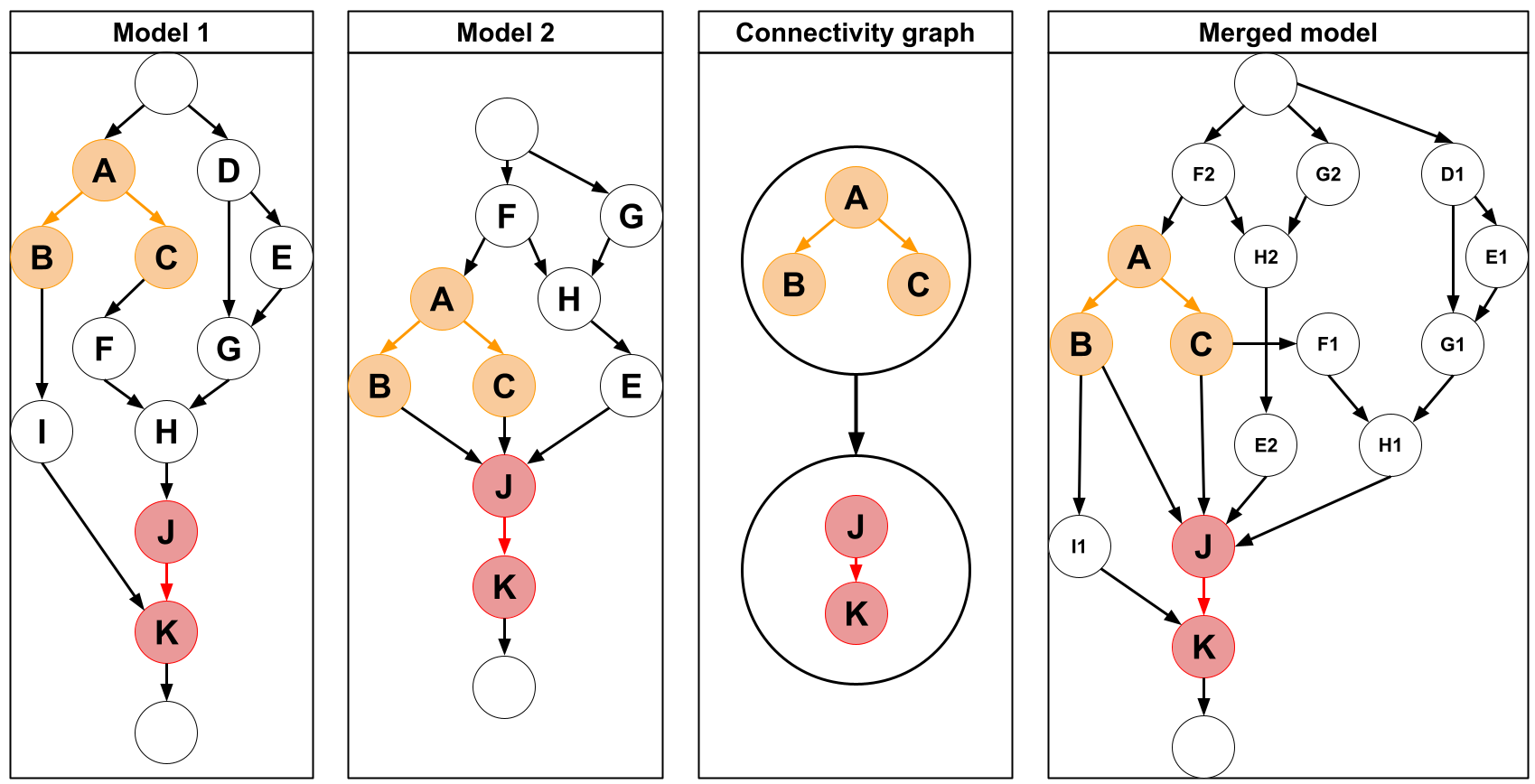}
    \caption{Merging DFG models with the naive approach}
    \label{fig:subgraph_as_node_merged}
\end{figure}

In our example,  two original acyclic DFGs are combined based on merging subgraphs corresponding to the vertices of the connectivity graph remaining after the cycle elimination procedure (see Fig.~\ref{fig:subgraph_as_node_merged}). 
Thus, common subgraphs that were excluded from the connectivity graph appear twice in the resulting model.
Existing methods and visualizations generally do not support DFG models with duplicate node names. Therefore, we add indices to duplicate names.

\subsubsection{More Accurate 
Merging}\label{sec:accurate}

In the naive approach, we duplicate the entire common subgraph if the corresponding vertex in the connectivity graph is excluded to break cycles (see Fig.~\ref{fig:subgraph_as_node}). 
   However, in many cases, it is possible to remove only some event nodes in common subgraphs and still break all cycles.

For a more accurate approach, we include entire subgraphs in the connectivity graph. 
  Now in the connectivity graph, an arc connects two vertices in two different common subgraphs iff there is a path from one of them to the other in at least one of the two initial acyclic DFGs.
  
\begin{figure}[htb]
    \centering
    \includegraphics[width=0.25\textwidth]{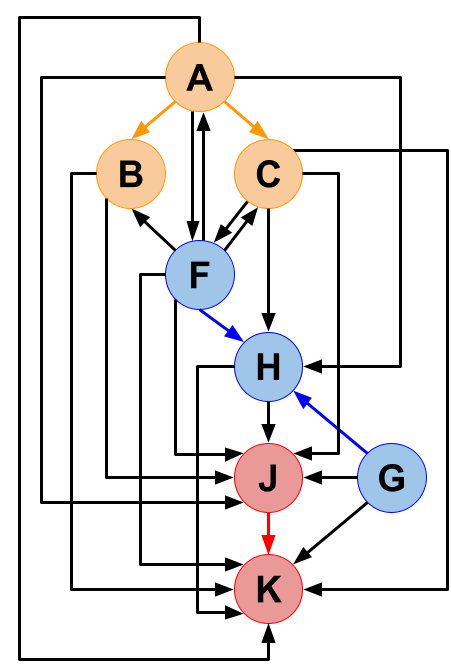}
    \caption{Connectivity graph with subgraphs for DFGs in Fig.~\ref{fig:common_subgraphs}}
    \label{fig:subgraph_full}
\end{figure}

We continue to use the example in Fig.~\ref{fig:common_subgraphs}, and  Fig.~\ref{fig:subgraph_full} shows the corresponding connectivity graph for accurate merging, where nodes of different subgraphs are coloured in different shades.

\begin{figure}[htb]
    \centering
    \includegraphics[width=0.5\textwidth]{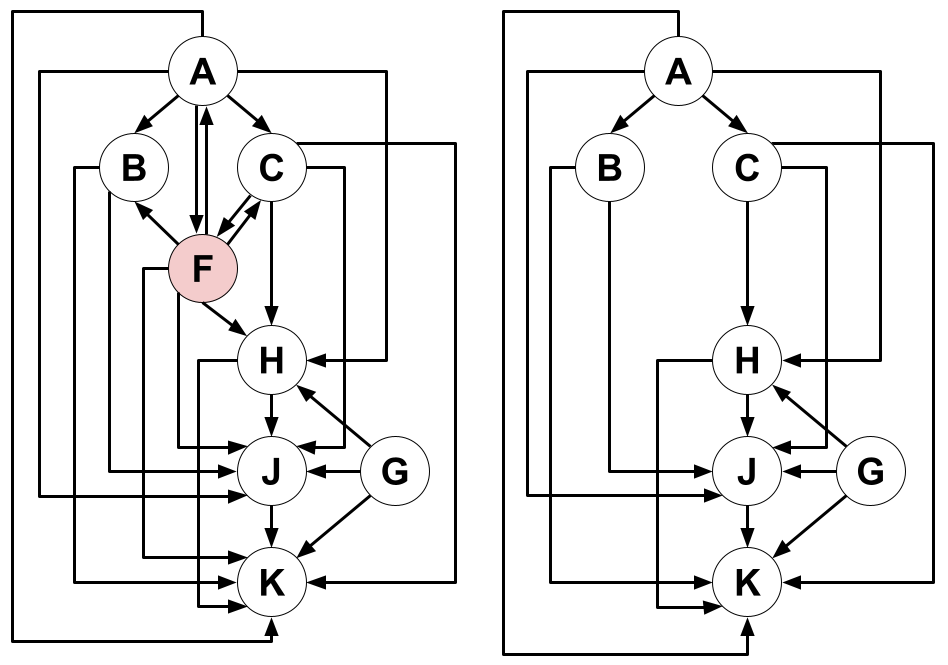}
    \caption{Node $F$ is removed to eliminate cycles in the connectivity graph}
    \label{fig:subgraph_fully_reduction}
\end{figure}

To eliminate cycles in the accurate approach, we use the same Bounded Search Tree algorithm as in the naive approach. In our example, we only need to remove one node $F$ to make the connectivity graph acyclic (see Fig.~\ref{fig:subgraph_fully_reduction}).

Compared with the naive approach, the accurate approach uses a connectivity graph with a larger number of nodes and arcs, so finding paths takes longer, as does finding the minimum FVS. Thus, the accurate approach takes longer.
However, the accurate approach produces a better model with a more compact structure and fewer duplicate names.

In our small example, to remove cycles, it is enough to remove only one node from the connectivity graph (Fig.~\ref{fig:subgraph_fully_reduction}), while the naive approach removes three vertices (Fig.~\ref{fig:subgraph_as_node_reduction}).

\begin{figure}[htb]
    \centering
    \includegraphics[width=0.9\textwidth]{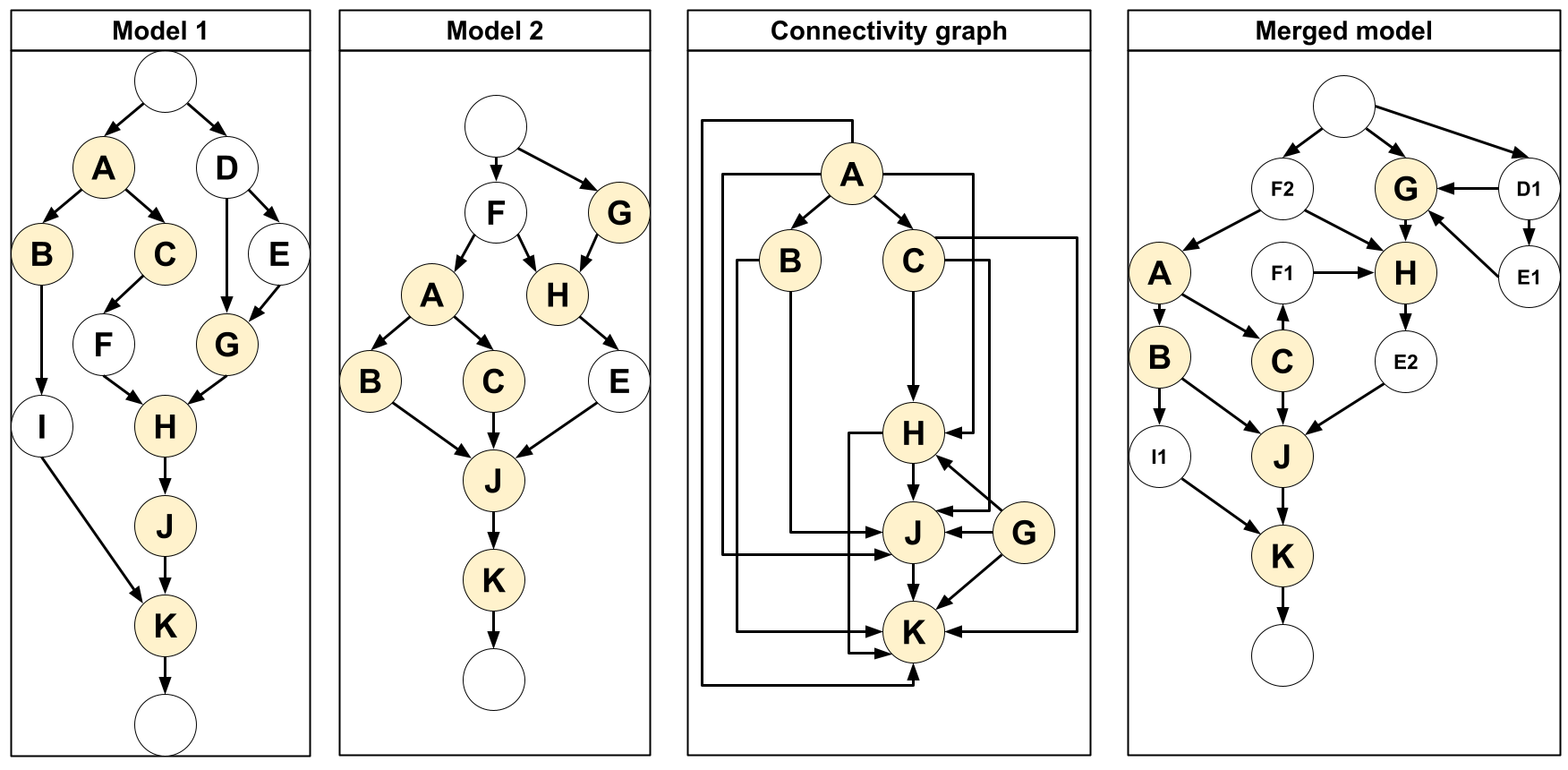}
    \caption{Merging models using the accurate approach}
    \label{fig:subgraph_fully_merged}
\end{figure}

Fig.~\ref{fig:subgraph_fully_merged} illustrates the complete DFG merging procedure using the accurate approach.

%------------------------------------------------------------------------------
\subsection{Merging two acyclic DFG models with duplicate event names}\label{sec:repeat}

The algorithms described above allow us to merge two models without duplicate event names, producing a DFG with duplicate names.
So, when we merge more than two models, to do it pairwise, we need an algorithm to merge acyclic models with duplicate event names. 

\begin{figure}[htb]
    \centering
    \includegraphics[width=0.35\textwidth]{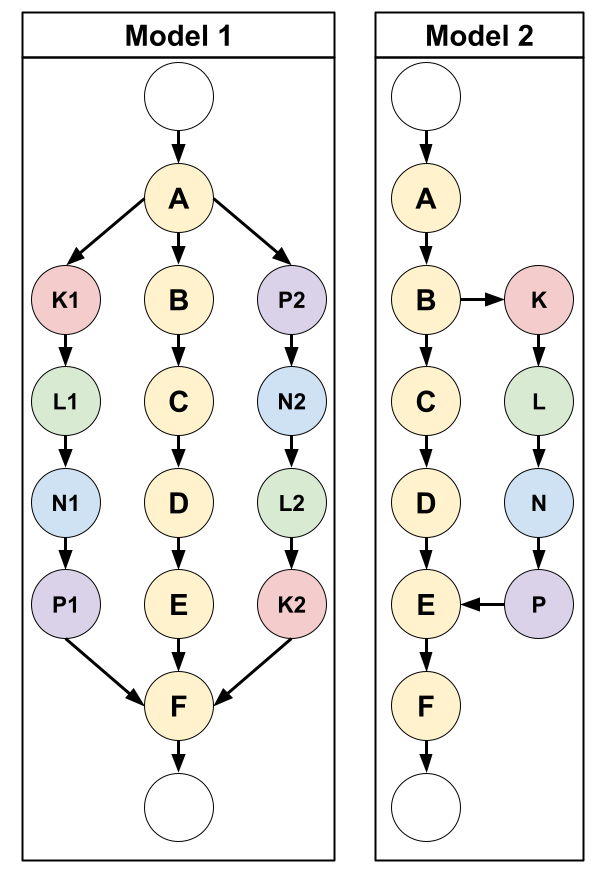}
    \caption{Merging models with repeated event names}
    \label{fig:repeating_merging}
\end{figure}

To illustrate the new algorithm, we use an example in Fig.~\ref{fig:repeating_merging}.
Here, two models have only one common subgraph with vertices $A$, $B$, $C$, $D$, $E$, and $F$. 
In the first model, event names $K$, $L$, $N$, and $P$ appear twice, but are renamed to distinguish them: $K$ into $K1$ and $K2$, etc. In the second model, each of these event names occurs once. 
The event name $K$ corresponds to the renamed event names $K1$ and $K2$, so when merging models, it can also be renamed to $K1$ or $K2$, or left unchanged. Thus, taking into account all the renaming options for each of the four nodes, we get 81 variants: $\{K, L, N, P\},\ \{K, L, N, P1\},\ \{K, L, N1, P1\},\ \dots$. 

To choose the best renaming option, we need to come up with a way to compare them. When merging models, our goal is to combine as many models as possible without creating cycles. Going through all the options would be too long. Taking into account the size of common subgraphs to determine the best option would be a faster approach, but at the cost of quality. The option chosen by this approach may not be the best one, since removing cycles may exclude some parts of the common subgraphs.

To reduce the number of combinations to check, we can rename nodes one by one. When evaluating each option, the arcs are taken into account, i.e. we search for common arcs. 
Therefore, we compare the number of common incoming and outgoing arcs for a node before and after the rename.
For example, if node $K$ was renamed to $K1$, then renaming node $L$ to $L1$ will add a new common arc from $K1$ to $L1$.

\begin{figure}[htb]
    \centering
    \includegraphics[width=0.65\textwidth]{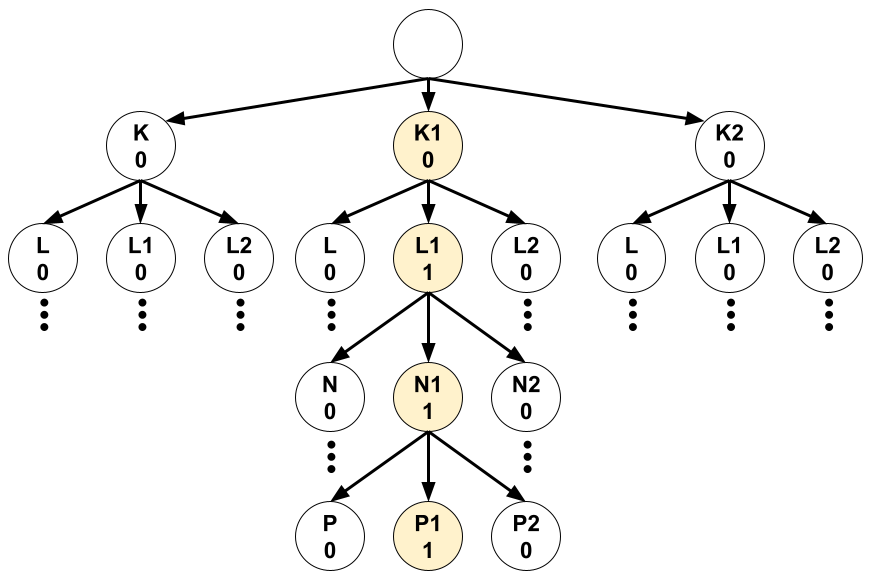}
    \caption{Tree with node renaming options}
    \label{fig:repeating_merging_tree}
\end{figure}

The process of renaming can be represented as a tree (see Fig.~\ref{fig:repeating_merging_tree}). A path from the root of the tree to a leaf represents one renaming option. Each node in the tree corresponds to one rename and is labelled with a number that indicates how many new common arcs will be created using this rename, taking into account past renames. The tree is built breadth-first and shows all possible renaming options.

We use a greedy algorithm to find an optimal path, which in our case defines a renaming option. The algorithm will select the node with the highest number and then continue to select among the child nodes of the selected node, and so on until it reaches a leaf. But the path may deviate towards the worst possible renaming options if the algorithm has to choose among the nodes that have only number 0. To improve results, we propose a set of empirical rules for the algorithm.

The first rule is applied when all the renaming options at the current depth level of the tree have number 0. In this case, the algorithm should choose among all child nodes of the current options. The parent of the chosen child node is also considered chosen. This way, more renaming options can be considered, and, in the worst case, all possible paths in the tree will be considered. But in turn, such an approach allows us to find better paths. In Fig.~\ref{fig:repeating_merging_tree}, at the level with node $K$ we have three zero options. Taking into account all child nodes at the next level, a single option $L1$ with $1$ is selected.

The second rule is applied when there are multiple choices with the highest number. In this case, the algorithm selects the same renaming label as the label of the parent node. For example, if at the first level the option $A1$ with a rename label $1$ was selected, and at the next level the rename options $B1$ and $B2$ both have the highest number, the algorithm selects the rename label of the parent node, i.\ e. option $B1$.

The order of the nodes in the tree significantly affects the result. The node number defines new common arcs after renaming and is calculated taking into account previously selected nodes. It may be that the order of the nodes in the tree results in a series of levels consisting of nodes numbered $0$. For example, if node $N$ was considered before node $L$, then at the first two levels of the tree, all nodes would have the number 0 (see Fig.~\ref{fig:repeating_merging_tree}). Such situations can increase the processing time and deviate the algorithm from better solutions. To avoid such cases, we arrange the nodes in the tree in approximately the same order as in the models.

Of course, when combining more than two models, the order of the models, due to the specifics of the proposed algorithm, can also significantly affect the results.

%------------------------------------------------------------------------------
\section{Evaluation}\label{sec:evaluation}

In this section, we evaluate our discovery approach using real-life data sets. 
The example provided in Section~\ref{sec:motivation} (see Tab.~\ref{tbl:example}) is a simplified sample extracted from the real-life event log that will be described in the following section.

\subsection{Real-life case description}\label{sec:data}

In this case study, we analyse educational trajectories of students who attend several university-level educational courses.
From the university LMS, we collected data on three groups of students who studied two educational courses over the course of half an academic year. 
In Tab.~\ref{tbl:data} shows an example of a sample of the data set. 
The grades in the data, depending on the class and course, may reflect either the student's grades, their attendance in classes, or whether they were active in class or not. To apply process discovery algorithms, the data must first be aggregated and converted into an event log in a suitable format.

\begin{table}[htb]
    \centering
    \caption{Example of data records}
    
    \begin{tabular}{c c c c} 
    \textbf{Student\_ID} & \textbf{Timestamp} & \textbf{Class} & \textbf{Grade} \\ \hline\hline
    \multicolumn{4}{c}{...}\\
    \hline
    1259066846049451744 & 06.09.2023 & lecture\_1 & 10 \\
    \hline
    1259066846049451744 & 08.09.2023 & seminar\_1 & 0 \\
    \hline
    1259066846049451744 & 13.09.2023 & lecture\_2 & 0 \\
    \hline
    1259066846049451744 & 15.09.2023 & seminar\_2 & 1 \\
    \hline
    \multicolumn{4}{c}{...}\\
    \hline
    1259066846049451744 & 06.12.2023 & lecture\_13 & 0 \\
    \hline
    1259066846049451744 & 08.12.2023 & seminar\_13 & 1 \\
    \hline
    1259066846049451744 & 13.12.2023 & lecture\_14 & 0 \\
    \hline
    1259066846049451744 & 15.12.2023 & seminar\_14 & 1 \\
    \hline
    1259066846049451744 & 21.12.2023 & midterm\_test & 2.56 \\
    \hline
    \multicolumn{4}{c}{...}\\
    \hline\hline
    \end{tabular}
  \label{tbl:data}
\end{table}

Based on the data obtained, a single unified event log was created (see Tab.~\ref{tbl:event_log} for an excerpt from this log). 
An event in this log is a class or an exam. 
Activities, which are also event names, are enhanced with additional data, to distinguish courses and grades. First, the name of the corresponding subject is added at the beginning of the event name. Next, at the end of the event name, we add the presence, activity, or level of the performance of the student: low, med[ium], high, or none. Some classes do not track student attendance and will only be marked as `none' or `active'.

\begin{table}[htb]
    \centering
    \caption{Example of records from the event log}
    
    \begin{tabular}{c c c c} 
    \textbf{Case\_ID} & \textbf{Timestamp} & \textbf{Activity} & \textbf{Grade} \\ \hline\hline
    \multicolumn{4}{c}{...}\\
    \hline
    1259066846049451744 & 2023-09-06 & bmgc lecture\_1 active & 10 \\
    \hline
    1259066846049451744 & 2023-09-08 & bmgc seminar\_1 absent & 0 \\
    \hline
    1259066846049451744 & 2023-09-13 & bmgc lecture\_2 none & 0 \\
    \hline
    1259066846049451744 & 2023-09-15 & bmgc seminar\_2 present & 1 \\
    \hline
    1259066846049451744 & 2023-09-15 & calc class\_1 present & 1 \\
    \hline
    \multicolumn{4}{c}{...}\\
    \hline
    -6531824385696264114 & 2023-09-04 & calc class\_1 present & 1 \\
    \hline
    -6531824385696264114 & 2023-09-06 & bmgc lecture\_1 none & 0 \\
    \hline
    -6531824385696264114 & 2023-09-08 & bmgc seminar\_1 absent & 0 \\
    \hline
    -6531824385696264114 & 2023-09-11 & calc class\_2 present & 1 \\
    \hline
    -6531824385696264114 & 2023-09-13 & bmgc lecture\_2 none & 0 \\
    \hline
    -6531824385696264114 & 2023-09-15 & bmgc seminar\_2 present & 1 \\
    \hline
    \multicolumn{4}{c}{...}\\
    \hline
    1259066846049451744 & 2023-12-18 & calc class\_14 present & 1 \\
    \hline
    1259066846049451744 & 2023-12-19 & calc homework med & 7.125 \\
    \hline
    1259066846049451744 & 2023-12-20 & calc quiz med & 3.5 \\
    \hline
    1259066846049451744 & 2023-12-21 & bmgc midterm\_test low & 2.56 \\
    \hline
    1259066846049451744 & 2023-12-21 & calc midterm\_test low & 2.4 \\
    \hline
    1259066846049451744 & 2023-12-22 & calc oral\_test high & 7.5 \\
    \hline
    \multicolumn{4}{c}{...}\\
    \hline
    -6531824385696264114 & 2023-12-15 & bmgc seminar\_13 absent & 0 \\
    \hline
    -6531824385696264114 & 2023-12-20 & calc homework low & 1.433 \\
    \hline
    -6531824385696264114 & 2023-12-21 & bmgc midterm\_test none & 0 \\
    \hline
    -6531824385696264114 & 2023-12-21 & calc quiz none & 0 \\
    \hline
    -6531824385696264114 & 2023-12-22 & calc midterm\_test none & 0 \\
    \hline
    -6531824385696264114 & 2023-12-25 & calc oral\_test none & 0 \\
    \hline
    \multicolumn{4}{c}{...}\\
    \hline\hline
    \end{tabular}
  \label{tbl:event_log}
\end{table}

Now, having the event log, we apply the process discovery algorithms and assess the discovered process models. To test our algorithm, we implemented the described approaches \citep{Repository} using the \emph{Python} programming environment and the \emph{NetworkX} library. 

\subsection{Models examples}\label{sec:models_examples}

\begin{figure}[htb]
    \centering
    \includegraphics[width=0.45\textwidth]{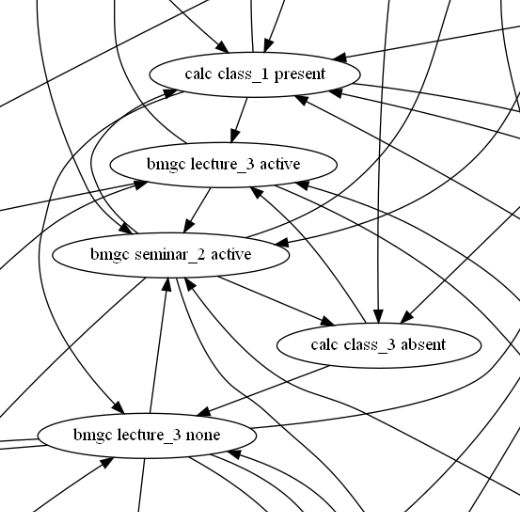}
    \caption{Fragment of the standard DFG model}
    \label{fig:cycles_example}
\end{figure}

As it was shown in the motivational example, the standard DFG discovery algorithm constructs the model with numerous cycles when dealing with the event log in which event names are ordered differently in different traces (see Fig.~\ref{fig:cycles_example}).
The resulting model looks quite confusing.
Such models are even called ``spaghetti models'' in informal conversations.
This type of model does not allow researchers to perform visual analysis due to its complexity. 

The main reason for the complexity of the model is the presence of cycles. Cycles occur because of differences in the order of event names. In our case, different groups of students have different dates for classes, as well as their numbers. This happens due to schedule mismatches between groups and various other circumstances. Thus, there are differences in the order and number of event names in the traces for different groups of students. 
The differences between groups grow with the number of educational courses involved as the number of possible mismatches increases. However, the nature of the process represented in this case is acyclic. 
In the real-life setting, students do not repeat exactly the same previous events, in our case, classes and exams. 
Thus, cycles should not be present in the model. 

\begin{figure}[htb]
    \centering
    \includegraphics[width=1\textwidth]{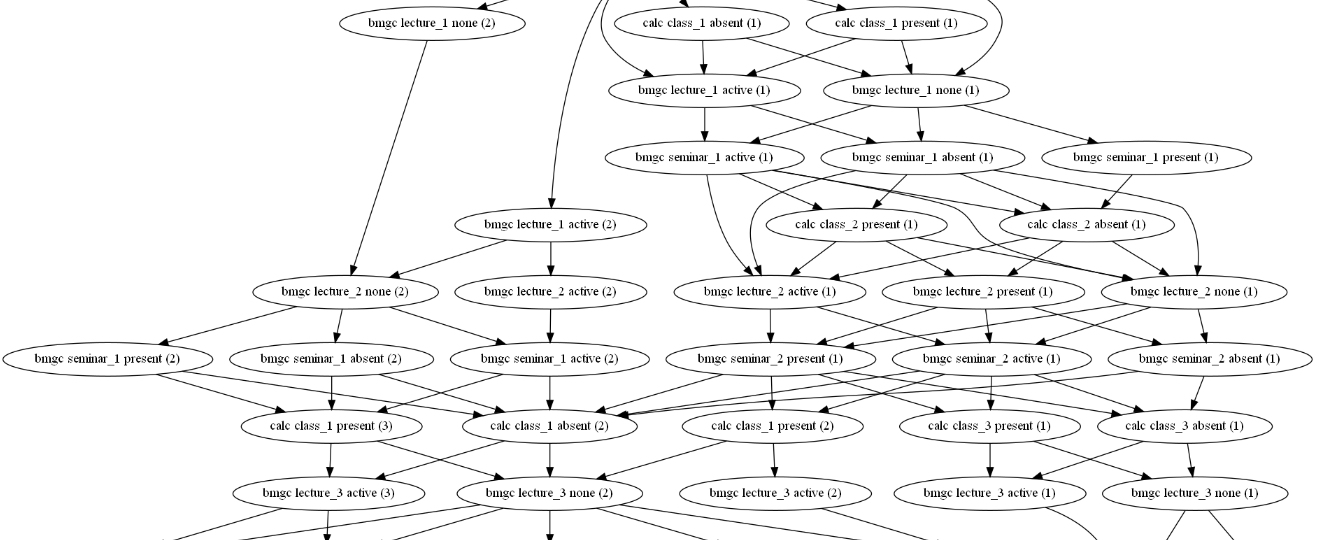}
    \caption{Fragment of the merged DFG model}
    \label{fig:merged_example}
\end{figure}

If we consider the process within a single group of students who share common dates of the classes, traces of students will have the same order of event names and dates of the events. Hence, such sequences of event names will not contain cycles. The considered acyclic event log can be partitioned into 3 parts so that each part contains one subgroup of students. Event logs for the subgroups are DFG-acyclic, and can be used to achieve acyclic DFG models. 
Then, these models can be merged using one of our approaches. 
As a result, we get a model that does not contain cycles (see Fig.~\ref{fig:merged_example}).

\subsection{Models assessment}\label{sec:models_assessment}

To assess discovered models and, thus, our algorithms, we calculate and compare various metrics for the process models, such as precision, fitness (recall) and time taken by the algorithms. 

We constructed several models using the acquired event logs for three student groups in two academic courses. As a standard DFG algorithm, we use the algorithm provided in the \emph{PM4Py} library \citep{BERTI2023100556}, with the results converted into a directed graph data structure from the \emph{NetworkX} library, for a more fair comparison with our algorithms built on top of this library. 
To test discovery through merging, we merge acyclic DFG models of individual student subgroups. By providing combinations with two of the three subgroups, more data can be obtained for testing.

Before comparing the metrics of the models, let us provide some statistical information about the discovered models, such as the number of records in the event logs used to synthesise models and the number of nodes in the models. To show the difference in cycles, we also provide the number of simple cycles in the models.
This statistical information is shown in Tab.~\ref{tbl:models_stats}.

\begin{table}[htb]
    \centering
    \caption{Models statistics}
    \begin{tabular}{l c l c c}
    \textbf{Event log} & \textbf{Event records} & \textbf{Model} & \textbf{Nodes} & \textbf{Simple cycles}  \\
    \hline\hline
    \multirow{3}*{All groups} & \multirow{3}*{3127} & Standard DFG & 127 & \textgreater 2 000 000 \\
    \cline{3-5}
    & & Merged DFG - naive approach & 224 & 0 \\
    \cline{3-5}
    & & Merged DFG - accurate approach & 224 & 0 \\
    \hline
    \multirow{3}*{Groups 1 \& 2} & \multirow{3}*{2328} & Standard DFG & 118 & 104 \\
    \cline{3-5}
    & & Merged DFG - naive approach & 165 & 0 \\
    \cline{3-5}
    & & Merged DFG - accurate approach & 165 & 0 \\
    \hline
    \multirow{3}*{Groups 2 \& 3} & \multirow{3}*{1877} & Standard DFG & 125 & 38 014 \\
    \cline{3-5}
    & & Merged DFG - naive approach & 201 & 0 \\
    \cline{3-5}
    & & Merged DFG - accurate approach & 201 & 0 \\
    \hline
    \multirow{3}*{Groups 1 \& 3} & \multirow{3}*{2049} & Standard DFG & 124 & 150 730 \\
    \cline{3-5}
    & & Merged DFG - naive approach & 201 & 0 \\
    \cline{3-5}
    & & Merged DFG - accurate approach & 201 & 0 \\
    \hline\hline
    \end{tabular}
  \label{tbl:models_stats}
\end{table}

The merged models contain about twice as many nodes as the DFGs discovered by the standard algorithm. 
As for cycles, there are none in merged models, whereas standard DFG models have a significant number of simple cycles. 
Again, it is important to note that the real process is acyclic and that all cycles in the model do not reflect the true behaviour. 
All traces present in the event log do not contain multiple instances of the same activity.

Now, to estimate how well our models correspond to the given data, we use standard metrics called precision and fitness (recall).  
Among various methods of conformance checking \citep{CarmonaDSW18}, there are two popular approaches to calculate fitness and precision: alignment-based replay \citep{Van_der_Aalst2012-kp} and token-based replay \citep{Rozinat2008-bd}. 
The \emph{PM4Py} library contains implementations of both these approaches to calculate fitness \citep{Berti2019-op, Buijs2014-dy} and precision \citep{Munoz-Gama2010-qe, Adriansyah2015-hv} based on Petri nets. \emph{PM4Py} also allows us to transform a DFG into a Petri net. 
Our models do not have any parallel events, so this transformation is safe.

Precision and fitness calculations for models with repeated events are a challenging procedure, because existing algorithms do not account for the possibility of multiple nodes associated with the same activity.  
During merging, if two event nodes are not combined into one node or if only one of the models contains an event node, the event names of these nodes will be renamed by adding a marker to their name. 
When calculating metrics, we also rename these event names in the corresponding event logs. So after merging, we combine the renamed event logs into a single event log and use it to calculate metrics for the merged model.

Another metric that we use is time. The execution time is measured with the \emph{Python} module \texttt{timeit}. 
We employ the same machine for all calculations.
Results are represented as mean time per loop and standard deviation over 7 runs with 100 loops each. The time measurements of the algorithms were made on a desktop PC with AMD~Ryzen~5 processor (6-Core~3.6~GHz) and 32~GB~DDR4~RAM. 

In fact, the application of the event log partitioning algorithm is not always necessary. For example, to obtain DFG-acyclic event logs in our case, we can manually divide the students by groups. In this regard, we additionally provide time measurements only for the merging algorithm, not taking into account the time required for the partitioning algorithm and construction of acyclic DFG models.

The measurement results are presented in one table (see Tab.~\ref{tbl:metrics}), which displays the fitness, precision, and time required for model construction. Both token-based and alignment-based replays provide almost the same precision and fitness across all of our experiments. Therefore, we do not specify the approach in the table.

\begin{table}[htb]
    \centering
    \caption{Model metrics}
    \begin{tabular}{lllll}
    \textbf{Event log} & \textbf{Model} & \textbf{Fitness} & \textbf{Precision} & \textbf{Time}  \\
    \hline
    \hline
    Group 1 & Standard DFG  & 1.0 & 1.0 & 6.27 ms ± 362 µs\\
    \hline
    Group 2 & Standard DFG & 1.0 & 1.0 & 5.82 ms ± 145\\
    \hline
    Group 3 & Standard DFG & 1.0 & 1.0 & 5.46 ms ± 123 µs\\
    \hline
    \multirow{7}*{All groups} & Standard DFG & 1.0 & 0.6049 & 8.5 ms ± 97.2 µs\\
    \cline{2-5}
    & Merged DFG - accurate approach & 1.0 & 0.8021 & \\
    & Only merging &  &  & 90.2 ms ± 1.08 ms\\
    & Full discovery &  &  & 172 ms ± 429 µs \\
    \cline{2-5}
    & Merged DFG - naive approach & 1.0 & 0.8021 & \\
    & Only merging &  &  & 73.5 ms ± 661 µs\\
    & Full discovery &  &  & 153 ms ± 2.53 ms\\
    \hline
    \multirow{7}*{Groups 1 \& 2} & Standard DFG & 1.0 & 0.5808 & 7.27 ms ± 87.5 µs\\
    \cline{2-5}
    & Merged DFG - accurate approach & 1.0 & 0.7551 & \\
    & Only merging &  &  & 25.2 ms ± 46.5 µs\\
    & Full discovery &  &  & 89.4 ms ± 1.14 ms\\
    \cline{2-5}
    & Merged DFG - naive approach & 1.0 & 0.7551 & \\
    & Only merging &  &  & 15.1 ms ± 144 µs\\
    & Full discovery &  &  & 77.9 ms ± 720 µs\\
    \hline
    \multirow{7}*{Groups 2 \& 3} & Standard DFG & 1.0 & 0.6998 & 6.73 ms ± 58.9 µs\\
    \cline{2-5}
    & Merged DFG - accurate approach & 1.0 & 0.7957 & \\
    & Only merging &  &  & 20.4 ms ± 439 µs\\
    & Full discovery &  &  & 72.7 ms ± 786 µs\\
    \cline{2-5}
    & Merged DFG - naive approach & 1.0 & 0.7957 & \\
    & Only merging &  &  & 14.1 ms ± 183 µs\\
    & Full discovery &  &  & 64.6 ms ± 767 µs\\
    \hline
    \multirow{7}*{Groups 1 \& 3} & Standard DFG & 1.0 & 0.6704 & 6.97 ms ± 44.8 µs\\
    \cline{2-5}
    & Merged DFG - accurate approach & 1.0 & 0.8265 & \\
    & Only merging &  &  & 14.5 ms ± 246 µs\\
    & Full discovery &  &  & 69.4 ms ± 288 µs\\
    \cline{2-5}
    & Merged DFG - naive approach & 1.0 & 0.8265 & \\
    & Only merging &  &  & 12.9 ms ± 195 µs\\
    & Full discovery &  &  & 68 ms ± 931 µs\\
    \hline
    \hline
    \end{tabular}
  \label{tbl:metrics}
\end{table}

From Tab.~\ref{tbl:metrics} we can see that merged models perfectly fit corresponding event logs and are more precise than corresponding standard DFG models. 
In our experiments, both naive and accurate approaches produce the same merged models. 

Considering the time required for discovering and merging, both approaches require more time than the standard DFG algorithm. This is not surprising: the amount of time required increases with the number of nodes in the common subgraph. 
The naive approach performs better in terms of time compared to the accurate approach. It is important to note that the time required to discover a single group model is excluded for merging, and we assume that these models are given. 

As was mentioned, for this real-life data, the models discovered using naive and accurate approaches are identical. But this is not always the case. In the following section, we show an example in which two approaches lead to different results.  

\subsection{Artificial data}\label{sec:artificial}

\begin{figure}[htb]
    \centering
    \includegraphics[width=0.49\textwidth]{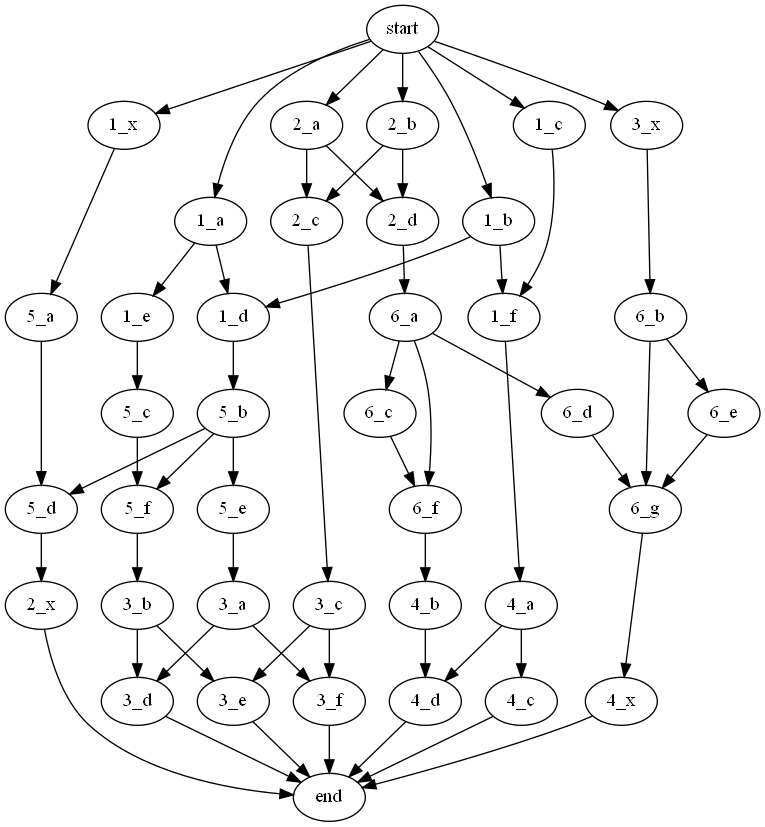}
    \includegraphics[width=0.49\textwidth]{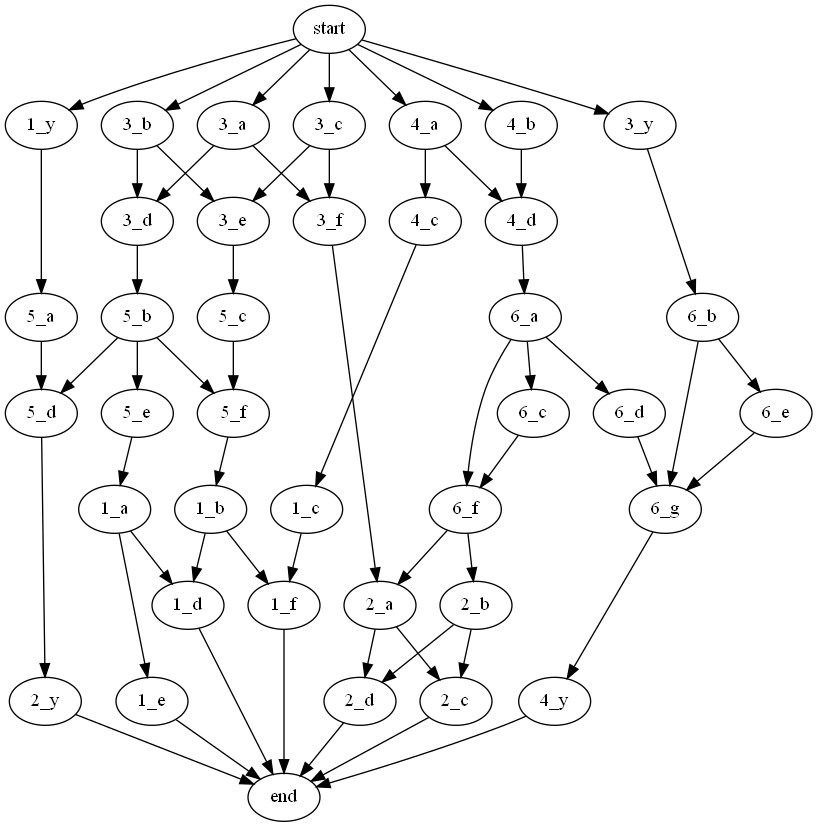}
    \caption{Process models for two departments}
    \label{fig:two_dep}
\end{figure}

To show a case where the accurate approach is preferable, let us provide an artificial one. In this case, we have two departments of an imaginable company that work on the same process. Each department has its own approach to the process (see the work models in two departments in Fig.~\ref{fig:two_dep}), so the order of the execution of activities is different.

The process model for each department is acyclic because event names are not repeated. Tab.~\ref{tbl:gen_models_stats} contains some statistics on the case. 
As in the previous example, simply combining event logs would lead to a DFG model with cycles. If a naive approach is used to build the model, there will be no merging of event nodes in this case. This model can be described as two acyclic DFG models for individual departments merged by the fusion of starting and ending nodes. 
Although the accurate approach in this case excludes only some parts of the common subgraphs, it allows more event nodes to be merged without creating cycles in the merged model.

\begin{table}[htb]
    \centering
    \caption{Artificial case statistics}
    \begin{tabular}{l c c c c}
    \textbf{Model} & \textbf{Nodes} & \textbf{Simple cycles} & \textbf{Fitness} & \textbf{Precision}  \\
    \hline\hline 
    Initial DFG for department 1 & 37 & 0 & 1.0 & 0.975 \\
    \hline
    Initial DFG for department 2 & 37 & 0 & 1.0 & 0.972 \\
    \hline\hline 
    Standard DFG for both departments & 41 & 27 & 1.0 & 0.975 \\
    \hline
    Merged DFG - naive approach & 74 & 0 & 1.0 & 0.985  \\
    \hline
    Merged DFG - accurate approach & 69 & 0 & 1.0 & 0.983 \\
    \hline\hline
    \end{tabular}
  \label{tbl:gen_models_stats}
\end{table}

Here we consider rather small example. However, it demonstrates the case where the accurate approach is more preferable for merging, and the difference between two approaches will grow with the size of the models. The accurate approach provides better merging when a process is broad and has many different execution paths, while different groups of process instances have significant differences in execution order. In less extreme cases, the naive approach provides solid results that take less time to compute.

%------------------------------------------------------------------------------
\section{Related work}\label{sec:related}

Process mining methods are covered by a vast number of works. 
Most aspects of process mining are covered in the books \citep{Aalst16,Burattin15,MansAV15} and in the subsequent handbook \citep{van2022process}. These works provide a solid foundation for the area with all the necessary definitions, methods, and algorithms.

The standard DFG discovery method is known to be problematic in some cases \citep{Van_der_Aalst2019-sp}. The work describes the problem with cycles that appear in process models containing parallel events or due to event names happening in different orders, even if in observable behaviour all event names appear at most once per trace. That implies that users should be aware of the features of the DFG discovery algorithm to avoid misleading interpretations.

BPMN models with repeated event nodes were considered in \cite{Lieben2018-st}, where the authors describe a discovery algorithm for exploratory data analysis. By allowing vertices to repeat in the models, the resulting models can be simplified. Repeating event nodes is also an alternative way to visualize parallel events. This creates less visual clutter and results in clearer and more accurate models.

The search for similarities and differences in process model patterns is present in many fields. Works that examine the area of concept drift focus on changes in the process over time \citep{Bose2011-hg, Bose2014-by}. This can also be described as finding the parts where the processes are the same and where they are different. Another area is the repair of process models \citep{FahlandA15,MitsyukLA17,PolyvyanyyAHW17}, where the goal is to improve the quality of a model by changing as few parts as possible. During the model repair process, the model is decomposed into parts \citep{Aalst13}, which are analysed separately. 
Bad parts can be repaired, and then all the parts are composed into a repaired model \citep{MitsyukLA17}. 
In addition, process models can be decomposed into parts to make them easier to compare more accurately \citep{BrockhoffGUA24}.

Assessment of process models is covered by conformance checking techniques \citep{CarmonaDSW18,Munoz-Gama16}. Two main ways to assess a model and an event log are alignments and token-based replay. The token-based replay approach was first mentioned in the work of \cite{Rozinat2008-bd}. The trace alignment approach was first mentioned in the work of \cite{Van_der_Aalst2012-kp}. With the help of these algorithms, we can obtain the precision and fitness of a model based on a given event log. Conformance checking algorithms also relate to the question of similarities and differences between process models and event logs. One of the works \citep{Artamonov2019-vq} described the conformance checking technique that can detect differences in model fragments through event relations. 

%------------------------------------------------------------------------------
\section{Conclusion and future work}\label{sec:conclusion}

In this paper, we propose a new approach to process discovery. This allows us to build acyclic DFG models for acyclic processes. Given an acyclic process, the acyclic event log can be partitioned into parts that produce acyclic DFG models, which can then be merged into a single acyclic DFG model representing the initial event log. 
The occurrence of cycles during merging is avoided by allowing the repetition of event nodes. The DFGs synthesised using the provided algorithm, compared with those constructed using the standard DFG discovery algorithm, retain the same level of fitness and provide a higher level of precision due to the absence of cycles.
However, the number of nodes in the discovered model is higher because of repeating nodes. 
The two merging approaches shown in this research provide a choice between speed and quality. The accurate approach requires more time to provide better merging in more complex cases. 
In contrast, the naive approach often returns the same result while taking less time to merge the models. We demonstrate how algorithms work
using a real-life example based on educational data from universities. Of course, the use cases of our approach are not limited to educational data or any other particular applied area.

The absence of cycles in the model opens up the possibility of using process model visualisations that would be useless and difficult to implement in the presence of cycles. 
One particular example of a cycle-dependent visualisation is the approach based on Sankey diagrams \cite{Derezovskiy}. 
Sankey diagrams allow us to show cycles, but cycles greatly reduce the clarity of the model. Another example where acyclic models are also required is Bayesian belief networks. In \cite{Vasilecas2014}, authors consider the extraction of a directed acyclic graph from an event log for the generation of Bayesian belief networks.

The main drawback of our discovery algorithm that requires improvement is its execution time. One possible way to improve its performance is to use a faster algorithm to find the minimum FVS. 

In terms of merging more than two models, a challenge is to find a better approach to merge DFG models with repeated nodes that would increase the merging speed for multiple models. The convenience of the algorithm can be further improved by adding automatic split of an event log into parts that provide acyclic DFG models, which can then be merged into a single acyclic model. 

In further research, the proposed approach can be applied to other modelling notations such as Petri nets or BPMN models, as well as to merging acyclic models with parallel events.

%------------------------------------------------------------------------------

\setstretch{2.0}
\bibliographystyle{user-spbasic}
\bibliography{references}

\end{document}